\RequirePackage{fix-cm}

\documentclass[twocolumn, natbib]{svjour3}

\smartqed  

\usepackage[ngerman,english]{babel}
\usepackage{amsmath}
\usepackage{graphicx}
\usepackage{booktabs}
\usepackage{amssymb}
\usepackage[colorlinks=true, linkcolor=blue, urlcolor=blue, citecolor=blue]{hyperref}

\journalname{PFG – Journal of Photogrammetry, Remote Sensing and Geoinformation Science}

\begin{document}

\title{PG-SAG: Parallel Gaussian Splatting for Fine-Grained Large-Scale Urban Buildings Reconstruction via Semantic-Aware Grouping}


\author{
    Tengfei Wang\textsuperscript{1} \and 
    Xin Wang\textsuperscript{1,*} \and 
    Yongmao Hou\textsuperscript{1} \and 
    Yiwei Xu\textsuperscript{1} \and 
    Wendi Zhang\textsuperscript{1} \and 
    Zongqian Zhan\textsuperscript{1}\thanks{* Corresponding author( \email{xwang@sgg.whu.edu.cn})}\thanks{SI on Sino-German P\&RS cooperation: Application, Methodology, Reviews}
}

\institute{
    \textsuperscript{1}School of Geodesy and Geomatics, Wuhan University, 129 Luoyu Road, Wuhan 430072, People’s Republic of China 
}
\date{Received: date / Accepted: date}


\maketitle

\begin{abstract}
3D Gaussian Splatting (3DGS) has emerged as a transformative method in the field of real-time novel synthesis. Based on 3DGS, recent advancements cope with large-scale scenes via spatial-based partition strategy to reduce video memory and optimization time costs. In this work, we introduce a parallel Gaussian splatting method, termed {\bf \textit{PG-SAG}}, which fully exploits semantic cues for both partitioning and Gaussian kernel optimization, enabling fine-grained building surface reconstruction of large-scale urban areas without downsampling the original image resolution. First, the Cross-modal model - Language Segment Anything is leveraged to segment building masks. Then,  the segmented building regions is grouped into sub-regions according to the visibility check across registered images. The Gaussian kernels for these sub-regions are optimized in parallel with masked pixels. In addition, the normal loss is re-formulated for the detected edges of masks to alleviate the ambiguities in normal vectors on edges. Finally, to improve the optimization of 3D Gaussians, we introduce a gradient-constrained balance-load loss that accounts for the complexity of the corresponding scenes, effectively minimizing the thread waiting time in the pixel-parallel rendering stage as well as the reconstruction lost. Extensive experiments are tested on various urban datasets, the results demonstrated the superior performance of our {\bf \textit{PG-SAG}} on building surface reconstruction, compared to several state-of-the-art 3DGS-based methods. Project Web: \url{https://github.com/TFWang-9527/PG-SAG}.

\end{abstract}

\begin{figure*}[htbp]
\centering
\includegraphics[width=1.0 \textwidth]{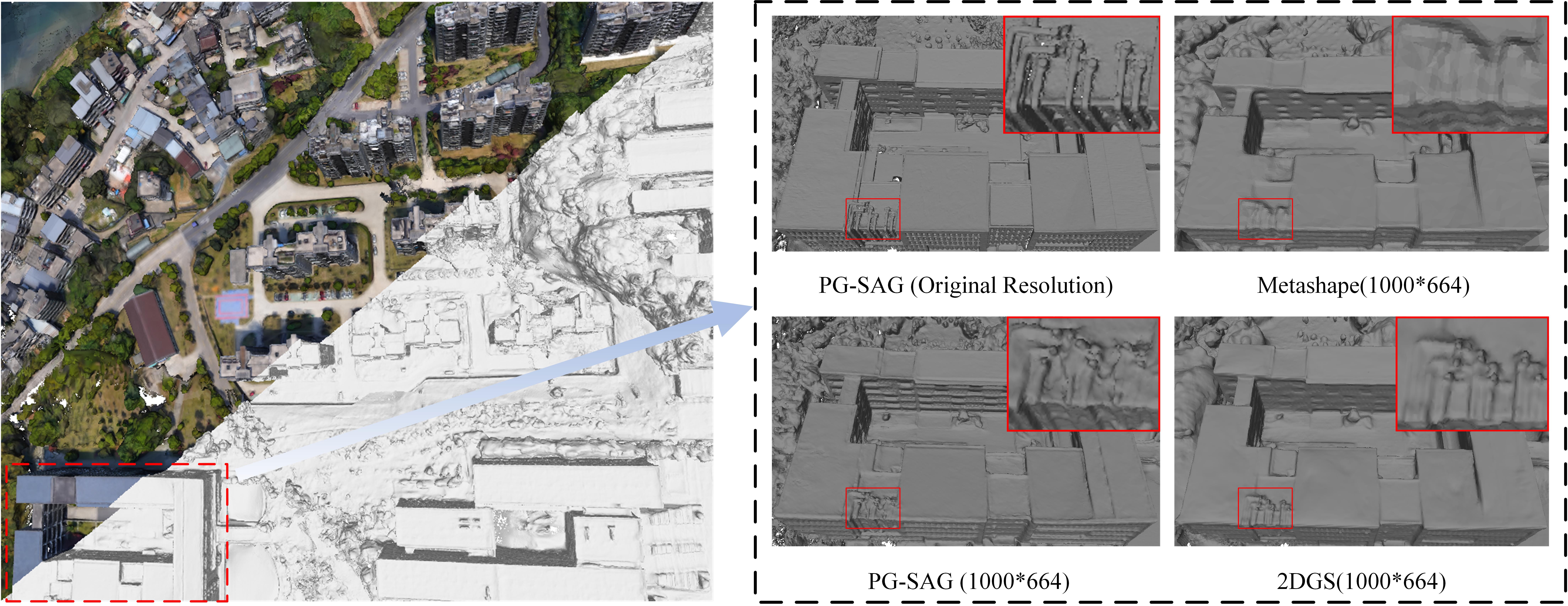}
\caption{Overall surface reconstruction results on the DPCV dataset, along with comparisons to other methods and our method using high-resolution images.Our PG-SAG with original resolution generates the most detailed meshes. Moreover, comparing to others of lower resolution, we again clearly perform better.}
\label{fig1}
\end{figure*}

\section{Introduction}
\label{intro}
Large-scale urban scene surface reconstruction has been extensively studied across various fields, such as photogrammetry \cite{3Dre_ph,3Dre_ph1,3Dre_ph2,3Dre_ph3}, computer vision \cite{3Dre_cv,3Dre_cv1,3Dre_cv2}. Among these studies, building reconstruction is a key focus and gains a wide range of applications, especially in road planning \cite{road} and digital city \cite{digitalcity}. Traditional approaches generally involve a series of complex processes, including structure from motion (SfM) \cite{SFM}, stereo dense matching \cite{stereo_dense_match} and multiple-view fusion  \cite{multi_view_fusion}, dense point cloud generation and filtering \cite{dense_point_filtering}, mesh generation \cite{mesh_gen_1,mesh_gen_2,mesh_gen_3}. In the last few years, NeRF \cite{nerf} have emerged as a promising implicit 3D scene representation, and its variants \cite{mega-nerf,block-nerf,mega-nerf++} have shown potential in large-scale scene novel view synthesis. The relevant NeRF-based surface reconstruction works \cite{Neus,Neus2,VOLSDF,geoneus} have also exhibited notable capabilities for object-level reconstruction; however, their prohibitive training costs limit their practical applicability to larger-scale scenes. More recently, newly developed 3DGS methods \cite{3DGS} have shown exceptional training efficiency and high fidelity in novel view synthesis tasks, sparking a wave of advancements in surface reconstruction leveraging 3DGS principles \cite{splatfields,3DGS,geoGaussian}.

\begin{figure}[htbp]
\centering
\includegraphics[width=0.5 \textwidth]{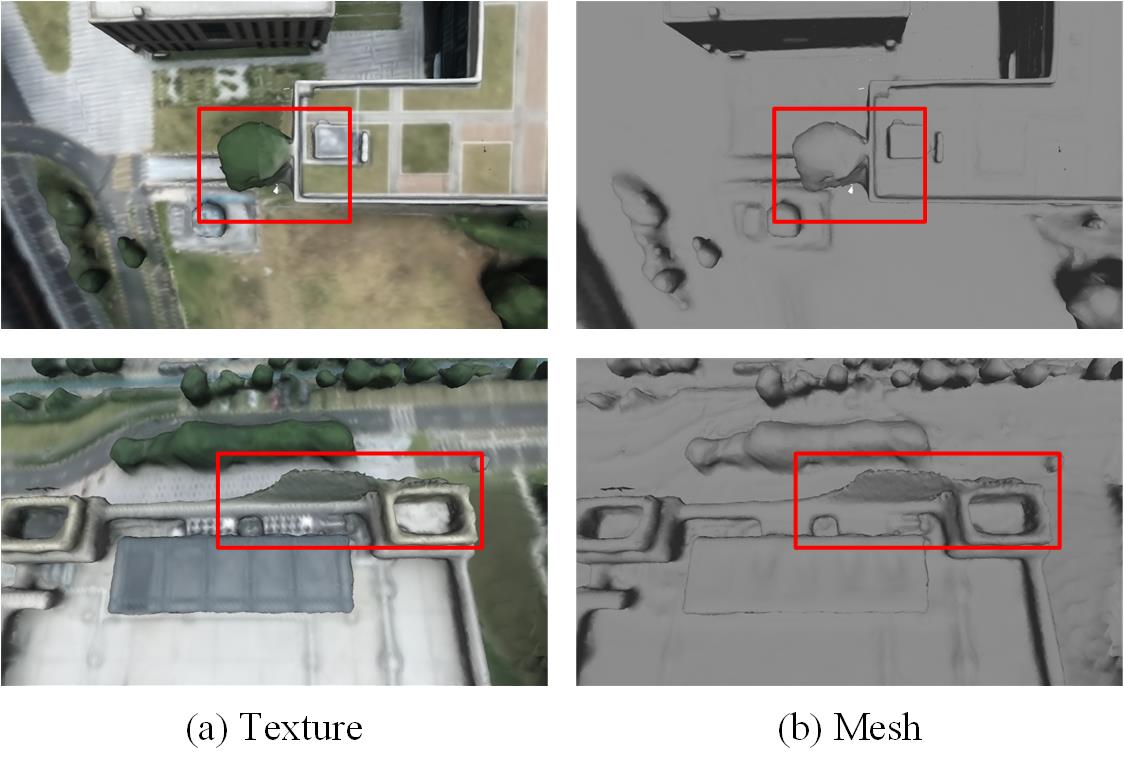}
\caption{Incorrect building meshes using PGSR\cite{pgsr}. Due to the interference from the background (non-building areas) on the foreground (building areas) when optimizing 3D Gaussians, erroneous reconstruction of building edges are produced, as shown by the highlighted details within the red boxes.}
\label{mix}
\end{figure}

Recent developments in 3DGS-based methods have demonstrated significant potential for high-quality surface reconstruction \cite{GOF,Sugar,pgsr,gs2mesh,localhint}. However, most of the methods remain confined to object-centric targets and small-scale scenes. When extended to large-scale urban scenes, several challenges are posed: \textit{First}, limited computational resources. Synchronously optimizing all Gaussian kernels for large-scale urban scenes is typically impractical due to computational resource constraints \cite{sags}. For instance, a single RTX 4090 GPU with 24GB can handle approximately 8.25 million 3D Gaussians, yet even a relatively small dataset like \textit{Garden} \cite{mip-nerf306}, covering less than 100$m^2$ already needs about 5.8 million 3D Guassians for high-fidelity rendering. One commonly adopted solution is to partition the large scene into multiple sub-blocks of smaller regions based on spatial locations, as seen in methods like Citygaussian \cite{citygaussian}, Vastgaussian \cite{vastgaussian}, and Gigagaussian \cite{Giga}, while simultaneously downsampling input images to accommodate memory requirement. However, as Fig.\ref{fig1} shows, the use of low-resolution images can hinder the precise pixel guidance required to optimize Gaussians, ultimately resulting in inaccurate reconstruction of the building surface. \textit{Second}, fine-grained building reconstruction. Buildings are among the most complex and highly scrutinized components in urban environments, making efficient fine-grained surface modeling particularly challenging. As Fig.\ref{mix} illustrates, the amalgamation and ambiguity of foreground (building) and background (non-building) during 3D Gaussian optimization can introduce noise, especially along edges \cite{MGF}, resulting in interference from non-building elements. This effect contributes to instability during training, particularly without sufficient iterations, and is even more pronounced in large-scale urban scenes.

To address these challenges, we propose \textbf{\textit{PG-SAG}}, the approach leveraging 3DGS for fine-grained building reconstruction within large-scale urban scenes. More specifically, to manage computational constraints, unlike \cite{citygaussian,  Giga,vastgaussian}, we introduce a semantic-aware grouping method to partition the large-scale urban scene.  Coarse masks of buildings are generated using Language Segment Anything (LSA) \cite{LSA}, followed by a reliability-scored multi-view voting filter that refines these masks to ensure multi-view consistency. The visibility among cameras and the correlation between cameras and sparse points are explored to group the building area of the entire scene into sub-groups,each encompassing its associated sparse points, cameras, and masks.  Each sub-group can be independently optimized in parallel. Notably, due to the use of masked pixels, our \textbf{\textit{PG-SAG}} accepts original high-resolution images directly without downsampling. In addition, the ambiguity of normal vectors along the building boundaries is addressed, recognizing that the referenced normal vector is not a real ground truth  at these boundaries. Thus, based on the boundaries derived from the detected masks, we reformulate the normal loss \cite{pgsr} into a boundary-aware normal loss, applying a balanced weight to the normal loss at the boundaries of the buildings. Lastly, to further reduce the training time caused by thread waiting during pixel-parallel rendering while minimizing the reconstruction lost, for each sub-group, we integrate a gradient-constrained balance-load loss, which take the complexity of the scenes into account. The more complexity scene typically yields more gradient information and needs a higher number of 3DGS for alpha blending, therefore, the balance load \cite{Adr-gaussian} is weighted by the constraints of gradient information.

In summary, our contributions are as follows:
\begin{itemize}
\item[$\bullet$] To the best of our knowledge, our \textit{\textbf{PG-SAG}} is the among the first 3DGS-based methods dedicated to fine-grained building reconstruction for large-scale urban areas.
\item[$\bullet$] We fully exploit the sematic cues for the partition of large-scale urban scenes and 3DGS optimization. Namely, a semantic-aware group partitioning strategy is proposed to address the limited computational resources, and the masked pixels are employed for 3D Gaussian optimization.
\item[$\bullet$] We present two improved losses, i.e., boundary-aware normal vector loss and gradient-constrained balance-load loss, to efficiently generate fine-grained building surface meshes of large-scale urban area.
\end{itemize}
\section{Related work}

\subsection{Multi-view Surface Reconstruction}
Over the last decades,  multi-view surface reconstruction has been a classical and fundamental topic in computer vision, computer graphics, and photogrammetry\cite{recon_bridge,recon_bridge2}. Traditional methods take the results of SfM as input, followed by multi-view stereo (MVS) methods that include voxel-based \cite{voxel_1}, surface evolution-based \cite{surface_evo_1}, depth-map-based approaches \cite{surface_evo_1}. These methods typically involve complex processes and are susceptible to matching errors\cite{mvs_match_error}. Subsequently, learning-based techniques have been explored to improve the performance of MVS and reconstruction\cite{mvs_deep_learning_1,mvs_deep_learning_2}; However,  the generalization might be limited when images from different domain are used\cite{unsup_mvs_deep_learning}.

NeRF and its variants \cite{Plenoxels,ISNGP,mip-nerf,NerfinW,mip-nerf306,nerf,nerf++} have been presented for novel view synthesis with an implicit encoding for the 3D scene, offering an alternative for surface reconstruction. NeuS \cite{Neus} and VolSDF \cite{VOLSDF} reformulate the inherent rendering volume as signed distance field (SDF)  to represent object surfaces and appearances, enabling smooth and accurate reconstructions of object-centric scenes. Neuralangelo \cite{Neuralangelo} further advances surface reconstruction by integrating the representational power of multi-resolution 3D hash grids with neural surface rendering. Despite their success,  these methods are predominantly suitable for small-scale scenes, and the substantial computational overhead associated with volumetric rendering poses challenges for scaling to large-scale scene reconstructions.

By leveraging 3D Gaussian primitives to explicitly model the appearance and geometry, 3DGS presents a considerable promise for surface reconstruction. One of the pioneering works, SuGaR \cite{Sugar}, generates meshes by constraining Gaussian spheres to align with surface features. 2DGS \cite{2DGS}  simplifies 3D Gaussians into directional 2D disks, improving both the geometric accuracy of the central target surface and computational efficiency.  GOF \cite{GOF} investigate the gaussian opacity fields, enhancing scene completeness in reconstruction compared to 2D Gaussian Splatting (2DGS). PGSR \cite{pgsr} enhances overall scene reconstruction accuracy by incorporating geometric and appearance consistency across multiple views. In addition, \cite{localhint}  trains neural implicit networks to approximate moving least squares (MLS) function in local regions, facilitating more accurate SDF from Gaussian spheres for mesh extraction.Unlike prior methods that directly extract scene geometry from Gaussian properties. GS2Mesh \cite{gs2mesh} derives geometry using a pre-trained stereo-matching model, mitigating the negative influence of noise originated from the individual depth profiles. While Gaussian-based methods have shown significant advantages in reconstruction speed and accuracy, their high memory demands limit the scalability for large-scale scene reconstructions\cite{vastgaussian,sags}.

\subsection{Large Scale Scene Reconstruction}
Comparing to object-centric or small-scale scenes, large-scale scene reconstruction normally needs to take into account time efficiency and memory constraints, where partitioning strategy and parallel processing techniques play crucial roles \cite{Rome,Rome_2}. Two popular traditional approaches have laid the groundwork in this area: \textit{Building Rome in a Day} \cite{Rome}, presented a novel distributed pipeline for parallel image matching and SfM, allowing the reconstruction of city-scale scenes. Similarly, \cite{Rome_2} decompose image collections into overlapping subsets that are processed in parallel to generate dense point clouds. These works enlighten subsequent large-scale scene reconstruction based on NeRF and 3DGS

Currently,  large-scale NeRF  and 3DGS focus primarily on novel view rendering tasks. For example, Block-NeRF \cite{block-nerf} and Mega-NeRF \cite{mega-nerf} adopt a divide-and-conquer strategy by partitioning scenes into smaller blocks,  with each block trained on a separate NeRF model. Grid-NeRF \cite{Grid-nerf} further integrates this strategy with feature grids, achieving notable improvements in rendering quality. In the domain of 3DGS, VastGaussian \cite{vastgaussian} employs a progressive partitioning and decoupled appearance modeling, which can reduce visual discrepancies during rendering. CityGaussian \cite{citygaussian} partitions the scene into blocks and introduces a Level-of-Detail (LoD) strategy, enabling fast rendering across multiple scales. In the context of large-scale scene 3D reconstruction, two concurrent related works explore 3DGS-based approaches. CityGaussianV2 \cite{cityGaussianv2} advances CityGaussian based on 2DGS, implementing  a decomposed-gradient-based densification and depth regression for eliminating floaters and expediting convergence. GigaGS \cite{Giga} concentrates on large scene surface reconstruction, it divides the scene based on mutual visibility of spatial regions, and multiview photometric and geometric consistency is explored to improve surface quality. 

\section{Preliminaries}
In our work, the 3DGS framework \cite{3DGS} is applied for optimizing 3D scene representation, leveraging the relevant unbiased depth rendering to estimate depth maps and obtain surface normal maps. In addition, to achieve fine-grained building surface reconstruction, we rely on the language segment-anything model (LSA) \cite{LSA} to mask building regions. To make this paper more self-contained, we shall provide an overview of the significance of these techniques as basics.

\subsection{3DGS and Unbiased Depth Rendering}
\textbf{3DGS}. 3DGS represents the 3D scene using a set of Gaussian spheres with multiple attributes, including spatial position, anisotropic variance, multi-order spherical harmonics to represent color, and transparency. For rendering, each 3D Gaussian sphere is transformed into a 2D Gaussian based on the viewing direction of each camera and then projected onto different image tiles. The 2D Gaussians are subsequently sorted and ${\alpha}$-blended to synthesis the output image:
\begin{equation}
  \textbf{\textit{C}} =  \sum_{i\in M} \textbf{c}_i\alpha_iT_j,   T_j=\prod_{j=1}^{i-1} (1-\alpha_j)
\end{equation}
all the encoded parameters are optimized by comparing to the referenced posed images using differentiable rendering.

\textbf{Unbiased Depth Rendering}. Inspired by 2DGS \cite{2DGS} and SuGaR \cite{Sugar}, to approximate the true surface, PGSR \cite{pgsr} flattens the 3D Gaussian ellipsoids into planes via minimizing the aligned scale factors, proposing an unbiased depth rendering method.The final normal map with current viewpoint is estimated via ${\alpha}$-blended:
\begin{equation}
  \textbf{\textit{N}} =  \sum_{i\in M} \textbf{\textit{R}}_c n_i \alpha_i T_i \label{eq:normal}
\end{equation}
where \textbf{\textit{$R_c$}} denotes the rotation from world system to the camera system, $n_i$ is the normal vector of the \textit{i}-th 3D Gaussian that passes through the ray which exhibits ambiguity at edges. The distance from the camera center to the flattened Gaussian plane is also rendered via ${\alpha}$-blended: 
\begin{equation}
  {\mathcal{\textit{\textbf{D}}}} =  \sum_{i\in M} d_i \alpha_i T_i
\end{equation}
in which, $d_i=(R_c^T(\mu_i-T_C))R_c^Tn_i^T$ is the distance from the camera center to the \textit{i}-th 3D Gaussian. Given the normal map and distances, PGSR calculates the corresponding depth map via the intersections between the rays and the planes, formulating depth that can precisely reflect the actual surfaces:
\begin{equation}
 \mathfrak{D}(p) =  \frac{\mathcal{\textit{\textbf{D}}}}{\textbf{\textit{N}}(p)K^{-1}\tilde{p}}
\end{equation}
where \textit{p} is the 2D position on the image, $\tilde{p}$ indicates the homogeneous coordinate of \textit{p}, and \textbf{\textit{K}} is the intrinsic matrix.

\subsection{Language Segment Anything}
In general, the LSA integrates two components of advanced models, GroundingDINO \cite{grounding} and SAM2 \cite{SAM2}. GroundingDINO is an open-set object detector that merges language and vision models, enabling object recognition within an image based on language prompts. SAM2, a versatile image and video segmentation model, supports multiple prompting and interaction modes. Driven by both visual and linguistic modalities, LSA yields superior segmentation performance that utilizes language prompts to segment images effectively, which enables efficient large-scale scene reconstruction with precise semantic segmentation.

\section{Method}
Large-scale scene reconstruction presents several key challenges, including limited computational resources and the need for fine-grained building surface mesh generation.  In this paper, we introduce PG-SAG, a novel approach that leverages 3D Gaussian Spheres to achieve detailed building reconstruction in large-scale urban scenes. Section \ref{SADG} elaborates the semantic-aware grouping approach, which partitions buildings with contextually relevant segmentation. Section \ref{BANL} details a boundary-aware normal vector loss to reduce edge ambiguities, and Section \ref{GCBL} introduces a gradient-constrained balanced-load loss. Finally, we explain the process of extracting and merging meshes.

\begin{figure*}[htbp]
\centering
\includegraphics[width=1.0\textwidth]{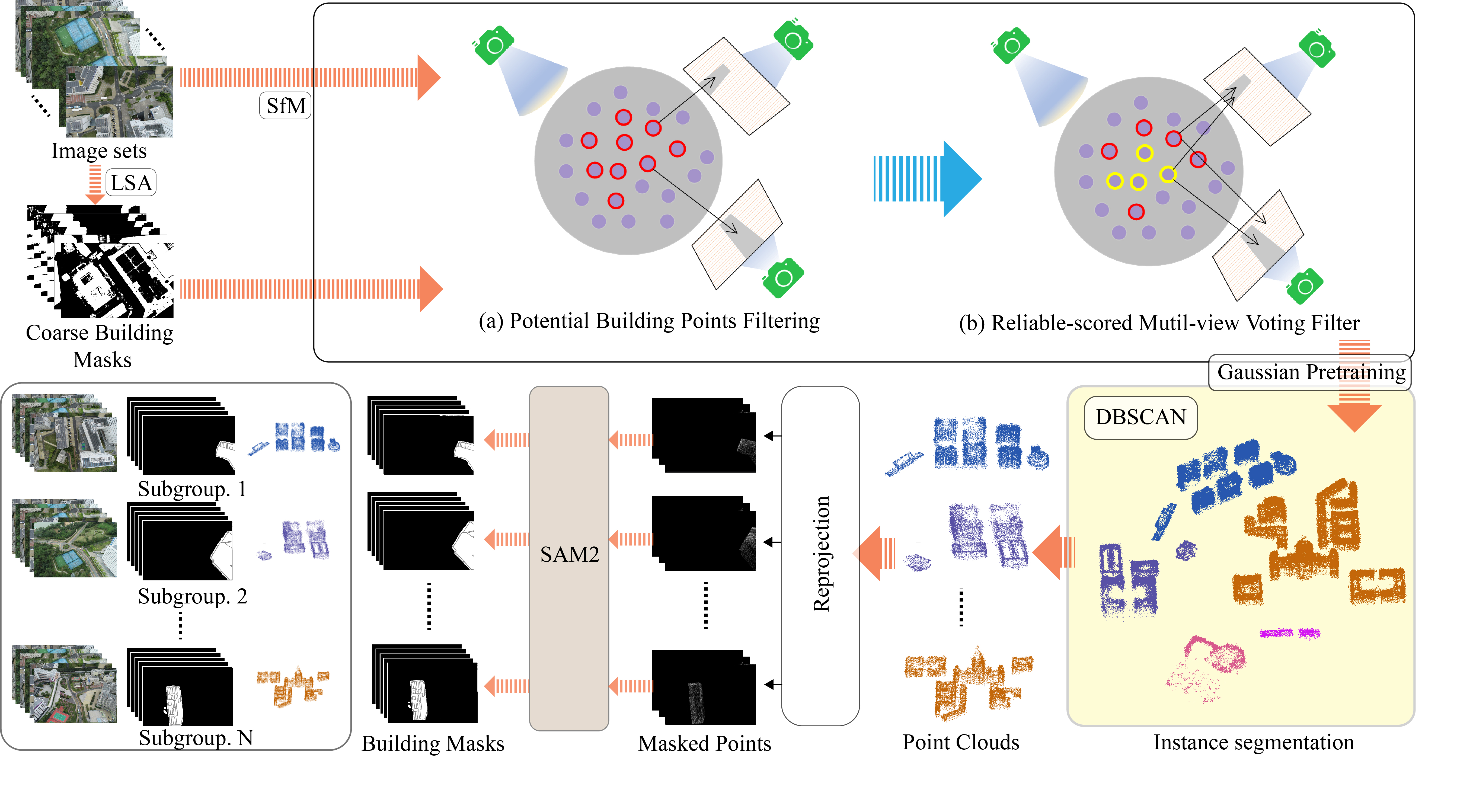}
\caption{Semantic-Aware Data Grouping Pipeline. The top-left part shows the coarse masks of buildings within the input images using LSA. The top-right parts illustrates a multi-view voting filtering, only points with high confidence, appearing in multiple building masks, are retained. The bottom part, from right to left, involves the usage of pre-trained Gaussian point-assisted point cloud instance segmentation, followed by reprojection to obtain mask points. In the final step, SAM2 is applied to extract refine building masks.}
\label{fig2}
\end{figure*}

\subsection{Semantic-Aware Data Grouping}\label{SADG}
We begin by apply the pre-trained LSA model to perform an initial, coarse segmentation of buildings and other objects within an image. Then, we refine these masks using a reliability-scored multi-view voting filter, enhancing segmentation accuracy and consistency across multiple views. Finally, we group the building regions of the entire scene into sub-groups based on camera visibility and the correlation between cameras and sparse points; each sub-group encompassing its associated sparse points, cameras, and refined masks. The pipeline of our semantic-aware data grouping strategy is shown in Fig.\ref{fig2}.

\textbf{Initial Segmentation. }
To generate initial building masks, we use LSA’s batch processing mode to segment multiple images synchronously, with text prompt "buildings" to predict the mask for building areas. At this stage, a relatively low promote threshold \textit{t} is selected to capture all building content as much as possible within the mask. Non-building areas are designated as "Background". Additionally, SAM2 \cite{SAM2} is applied to obtain a fine-grained segmentation mask for the entire image, covering both building and non-building regions.

Let $I$ denote the input image, and $p_{\text{building}}$ represent the text prompt "buildings". The initial building mask $M_{\text{building}}$ is derived by setting a threshold $t$:
\begin{equation}
M_{\text{building}} = \text{LSA}(I, p_{\text{building}}, t)
\end{equation}
where a small value $t$ is typically chosen to ensure that the mask $M_{\text{building}}$ includes all regions potentially associated with buildings.

\textbf{Reliability-scored Multi-view Voting Filter. }
To address initial coarse masks that stemmed from extraneous and false predicted building pixels, we propose a heuristic reliability-scored multi-view voting filter to refine the masks and achieve consistent, precise building masks across multiple views. First, the SfM sparse points are re-projected to each coarsely masked image, and for each point, we then count the frequency it locates in a masked region across all images. Point \( p_i \)  with the frequency value greater than zero is identified as a potential building point \textbf{\textit{\( p_p \)}}: 
\begin{equation}
\{p_p = p_i \mid \sum_{j=1}^{N} \mathbb{I}(p_i \in M_j) > 0 \}
\end{equation}
where, \( N \) is the number of input images, \( M_j \) represents the mask for the \( j \)-th image, \( \mathbb{I}(p_i \in M_j) \) is an indicator function that returns 1 if \( p_i \) lies within the masked region \( M_j \) of the \( j \)-th image, and 0, otherwise.

For each potential building point, we calculate an "unreliability score ($\textit{US}$)", defined as the number of times the point appears in the image but outside the corresponding mask. A potential building point is considered as a reliable building point if its $\textit{US}$ is below a predefined tolerance ($\tau$), which is empirically selected depending on the degree of image overlapping information. 
\begin{equation}
\text{\textit{US}}(p_p) = \sum_{j=1}^{N} \mathbb{I}(p_p \in \text{Image}_j \setminus M_j)
\end{equation}
where \( \text{Image}_j \setminus M_j \) represents regions outside the mask \( M_j \) in the \( j \)-th image, \( \mathbb{I} \) is an indicator function that equals 1 if \( p_i \) lies within \( \text{Image}_j \setminus M_j \), and 0 otherwise.

\begin{figure}[htbp]
\centering
\includegraphics[width=0.5\textwidth]{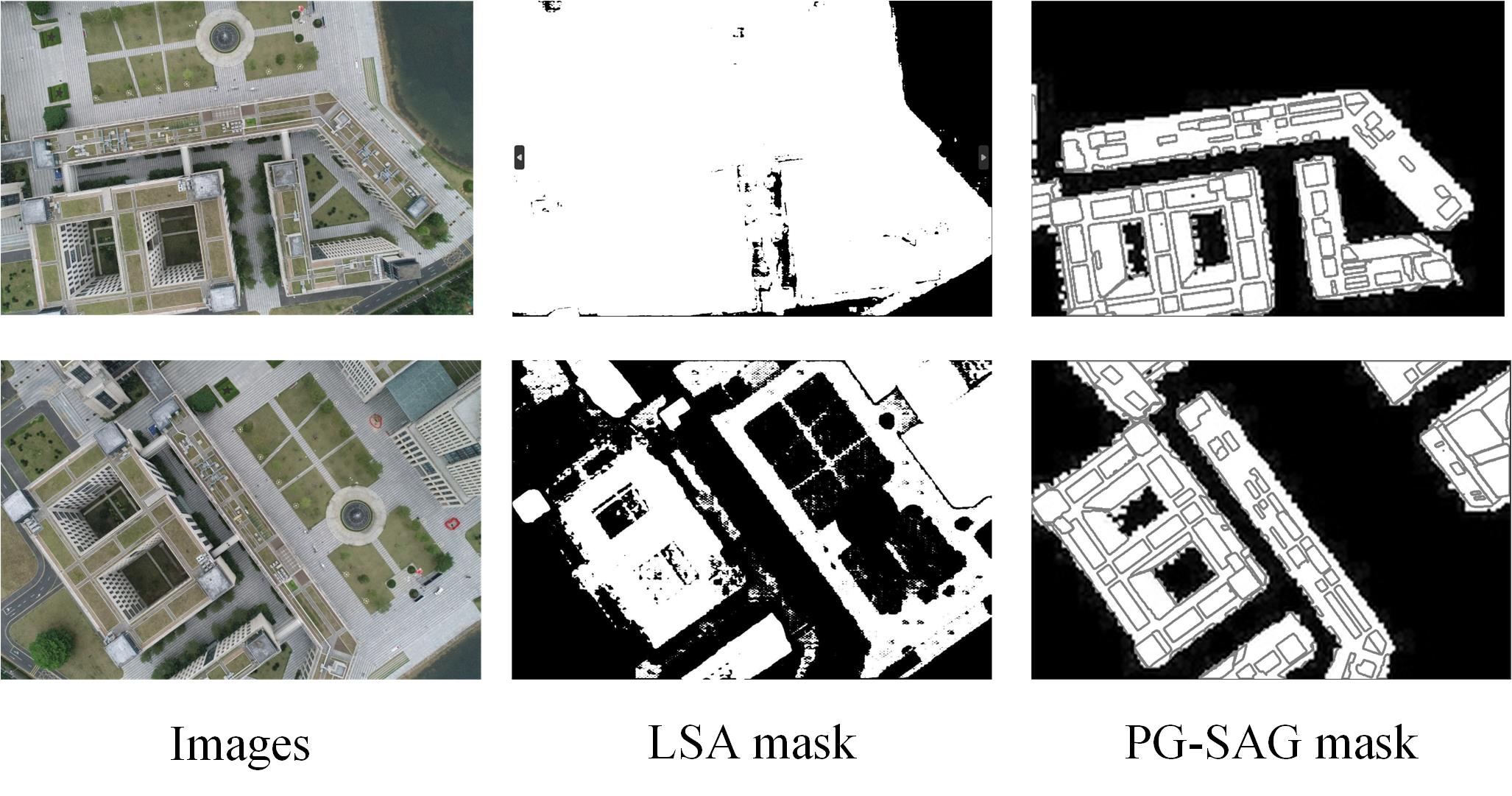}
\caption{Comparison of different segmentation methods. LSA (lang-segment-anything) confuses the ground with buildings, resulting in inaccurate masks. PG-SAG can not only obtain complete building masks, but also obtain fine boundaries.}
\label{mask}
\end{figure}

After determining reliable building points, we extend them using pre-trained Gaussian points which are projected onto the fine-grained segmentation mask as point-based queries. Then, based on SAM2, we generate a new refined mask with accurate building boundaries $\textbf{\textit{RBM}}$, as well as the edges of the internal building object. Some segmentation comparison are shown in the Fig.\ref{mask}. 

\textbf{Building Grouping. }
We use the DBSCAN algorithm \cite{DBSCN} to group building regions based on geographic proximity, clustering nearby buildings into subgroups that each contain the corresponding sparse point cloud, original images, and already refined masks (see Fig.\ref{fig3} for an example). These subgroups are then independently optimized in parallel with the original image resolution. For non-building objects, we use a similar geographic partitioning approach, as in VastGaussian, to divide areas outside the building masks into geographic blocks.

\subsection{Boundary-Aware Normal Loss}\label{BANL}
The original 3DGS optimizes the Gaussian kernels via photometric image reconstruction, but it often falls into local overfitting minima, causing 3D Gaussian to misalign with the actual surface.  To address this, the PGSR \cite{pgsr} extends 3DGS by incorporating the Local Plane Assumption, which enforces local consistency in depth and normals by approximating each pixel and its neighbors as a planar region. This assumption handles non-local planarity at the edges through the gradient-base regularization. 

However, as Fig.\ref{fig5} shows,  the Local Plane Assumption struggles in fine-grained building surfaces. While effective on flat areas, it deteriorates near building edges, where sharp depth changes and uncertain normal vectors occur (see equation \eqref{eq:normal}). Thus, we leverage the refined masks to extract more accurate building masked boundaries, noted as $\textbf{\textit{MB}}$,  and propose a \textbf{ boundary-aware normal loss}. Specifically, in computing the loss between the normals derived from depth map and rendered normals, we apply adaptive weighting at masked boundaries to reduce the ambiguous impact of normal loss.

Let \( \mathbf{\textit{\textbf{n}}}_{\text{\textit{\textbf{depth}}}} \) denote the normal vector computed from the depth map using four neighboring points, and \( \mathbf{n}_{\text{rendered}} \) denote the normal vector obtained from the rendering process. The boundary-aware normal loss \( L_{\text{ban}} \) is defined as:

\begin{equation}
\mathcal{L}_{\text{ban}} = \sum_{i} w_i \|\mathbf{n}_{\text{depth}, i} - \mathbf{n}_{\text{rendered}, i}\|^2
\end{equation}
where \( w_i \) is a weight factor that varies depending on whether pixel \( i \) lies on the building boundary. Specifically, for $pixel(i)\in \textit{\textbf{RBM}}$, \( w_i \) is assigned a small value OF 0.1 to reduce the influence of normal ambiguity in these regions/ Otherwise,  \( w_i =1\).

\subsection{Gradient-constrained Balance-load Loss}\label{GCBL}

The 3DGS performs point-based rendering, where each pixel's color is calculated in parallel through Gaussian rasterization, with each pixel mapped to a distinct GPU thread. However,  load imbalances arise due to different number of Gaussians across pixels and limits the efficiency. To further reduce training time within each subgroup by minimizing thread idle periods during parallel pixel rendering, referring to AdR-Gaussian \cite{Adr-gaussian}, the balance-load loss is advertised. Basically,  this loss constrains the number of Gaussians per pixel, promoting consistent workload distribution across threads. Nonetheless, directly enforcing this loss risks compromising reconstruction fidelity, as more intricate scenes typically generate additional gradient information and necessitate a higher density of 3D Gaussians and Splatting operation to support better ${\alpha}$-blending. 

Therefore, we propose a \textbf{Gradient-Constrained Load Balancing Loss} that accounts for scene complexity to balance Gaussian distribution while preserving reconstruction quality. This approach aims to expedite training by adapting the number of Gaussian spheres according to scene complexity. The \textbf{Gradient-Constrained Load Balancing Loss} \( L_{\text{GC-load}} \) is defined as follows:
\begin{equation}
\mathcal{L}_{\text{GC-load}} = \operatorname{\textit{std}}_{i \in HW} (g_i/w_i),
\end{equation}
where \( w_i \) is a gradient-dependent weight $\nabla I$ for pixel \( i \), \( g_i \) denotes the number of Gaussian spheres contributing to pixel \( i \), \(\operatorname{\textit{std}}\) represents the standard deviation over all pixels the grid of \( H \times W \). Higher gradients indicate a more complex scene, which allows for a higher variance in the number of Gaussian spheres to reduce reconstruction loss.

For 3DGS optimization, we adopt the multi-view consistency geometric loss and photometric consistency loss from PGSR, denoted as \( \mathcal{L}_{\text{mvgeo}} \), \( \mathcal{L}_{\text{mvrgb}} \), and the flattening 3D Gaussian loss \( \mathcal{L}_s \). But, only the refined building masks $\textbf{\textit{RBM}}$ are involved in these losses. In addition,  our boundary-aware normal loss \( \mathcal{L}_{\text{ban}} \). is also incorporated. We define the geometric loss \( \mathcal{L}_{\text{PG-geo}} \) as follows:
\begin{equation}
\mathcal{L}_{\text{PG-geo}} = \mathcal{L}_\text{\text{rgb}} + \lambda_1 \mathcal{L}_{\text{mvgeo}} + \lambda_2 \mathcal{L}_{\text{mvrgb}} + \lambda_3 \mathcal{L}_s + \lambda_4 \mathcal{L}_{\text{ban}}
\end{equation}
Then, the overall loss \( L \) is defined as:
\begin{equation}
\mathcal{L} = (1 - \lambda) \mathcal{L}_{\text{PG-geo}} + \lambda \mathcal{L}_{\text{GC-load}}
\end{equation}
where the weighting coefficients are set as follows: \(\lambda = 0.41\), \(\lambda_1 = 0.05\), \(\lambda_2 = 0.2\), \(\lambda_3 = 100\), and \(\lambda_4 = 0.01\).

\subsection{Mesh Extraction and Merge}
For the buildings of large-scale scene, by incorporating with our PG-SAG, a parallel group-based training solution is adopted to efficiently facilitate the building mesh generation from the optimized Gaussians. While the background regions are trained using a partition-based approach similar to vast-Gaussian. Subsequently, we generate depth maps for both buildings and background from multiple perspectives. For overlapping regions within the depth maps, mean fusion is applied, followed by mesh extraction from the depth map using the Truncated Signed Distance Function (TSDF) method \cite{TSDF}.

\section{Experiments}
\subsection{Experimental Setup}
\begin{figure*}[htbp]
\centering
\includegraphics[width=1.0\textwidth]{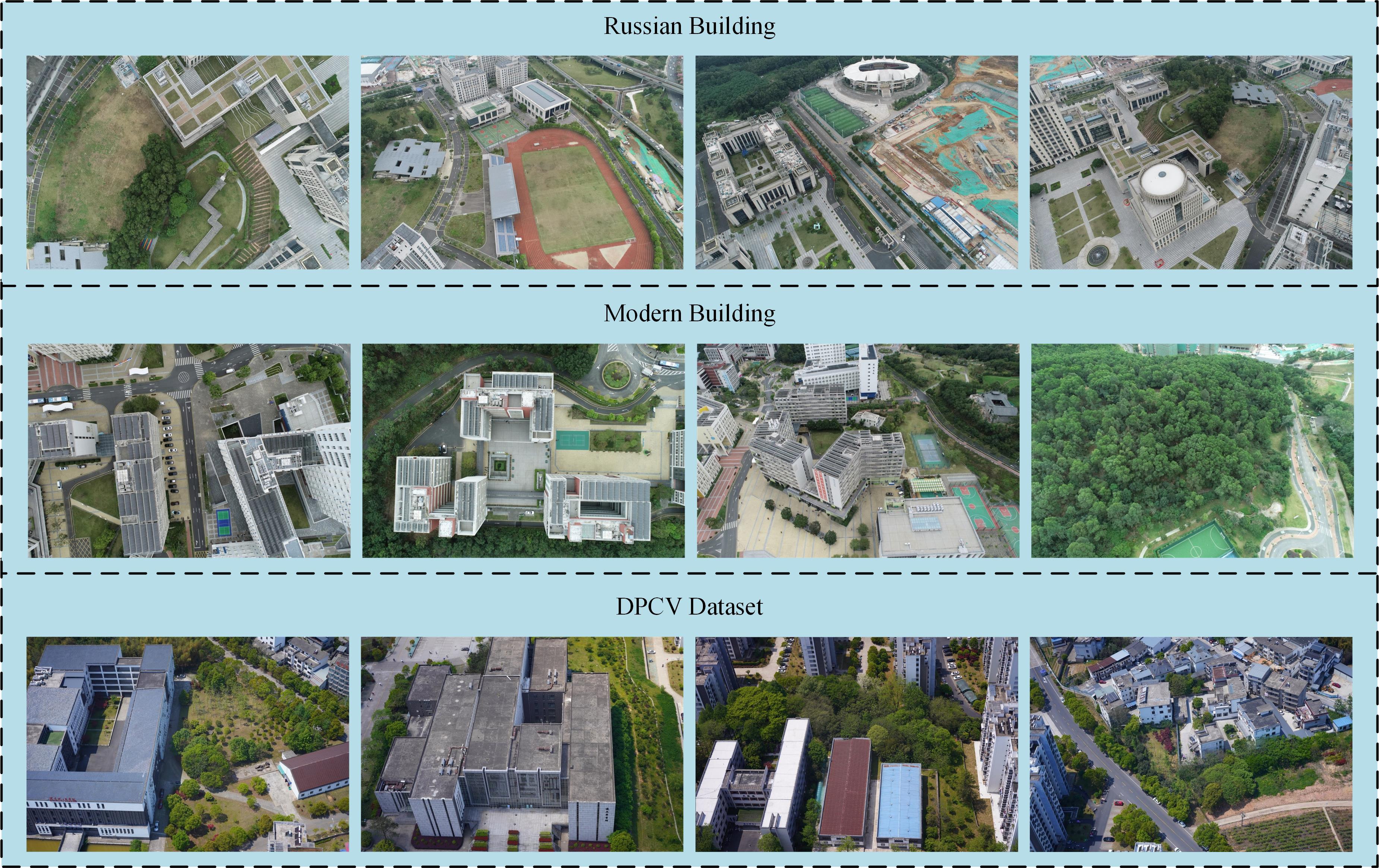}
\caption{Sample images from Russian Building, Modern Building and DPCV Dataset.}
\label{sample_images}
\end{figure*}

\textbf{Datasets and Metrics}
To validate the performance of our PG-SAG, particularly on the buildings in large-scale scenes, the GauU-Scene dataset \cite{GauU} is employed, which provides ground-truth point clouds for quantitative comparison. Specifically, we select the Russian Building and Modern Building scenes for both geometric evaluation and qualitative analysis. These two scenes consist of 713 and 563 images, respectively, and feature a variety of building types with a resolution of 5468 × 3636. Additionally, we conduct further qualitative analysis using the self-generated DPCV dataset collected from Huangshan China, which comprises 735 images with a resolution of 4745 × 3164. The flight height is 120 meters, covering an area of 1.7×1.7 $\text{km}^2$. The sample images are shown in Fig.\ref{sample_images}

For the geometric evaluation metrics of the mesh, we follow the approach outlined in \cite{TnT,Nerf3D}. First, surface sampling is performed on the reconstructed mesh, followed by the calculation of the corresponding F1 score, precision, and recall. To ensure a fair comparison and account for memory limitations, the maximum edge length of all images is downsampled to 1000 pixels.

\textbf{Baselines and Implementation.}
We compare our method with several state-of-the-art geometric reconstruction methods, including the 3DGS-based methods GOF, 2DGS, and SuGaR, as well as the advanced traditional photogrammetry software Metashape\cite{metashape}. All methods, including ours, are executed on four RTX 4090 GPUs. For the data partitioning, following the approach of VastGaussian, all datasets are divided into a 3×3 grids. In the qualitative evaluation, each block is trained for 30,000 iterations, with sparse point clouds and camera poses generated using COLMAP \cite{colmap}, and default parameters are used for all methods. In the quantitative evaluation, we align with the settings of CityGaussianV2 for comparison. While CityGaussianV2 uses 60,000 iterations for training, we maintain a training duration of 30,000 iterations due to the advantages of our data partitioning strategy.

 \begin{table}[htbp]
\centering
\caption{Performance Comparison on GauU-Scene.The best results are highlighed in bold}
\label{table:Gauu_scene}
\begin{tabular}{lcccc}
\toprule
Methods  & Precision $\uparrow$ & Recall $\uparrow$ & F1 $\uparrow$ \\
\midrule
GOF & 0.370 & 0.290 & 0.374 \\
2DGS & 0.553 & 0.446 & 0.491 \\
SuGaR & 0.570 & 0.292 & 0.377 \\
Metashape & 0.604 & 0.368 & 0.458 \\
PG-SAG & \textbf{0.671} & \textbf{0.467} & \textbf{0.551} \\
\bottomrule
\end{tabular}
\end{table}

\subsection{Comparison with SOTA Methods}

\begin{figure*}[htbp]
\centering
\includegraphics[width=1.0\textwidth]{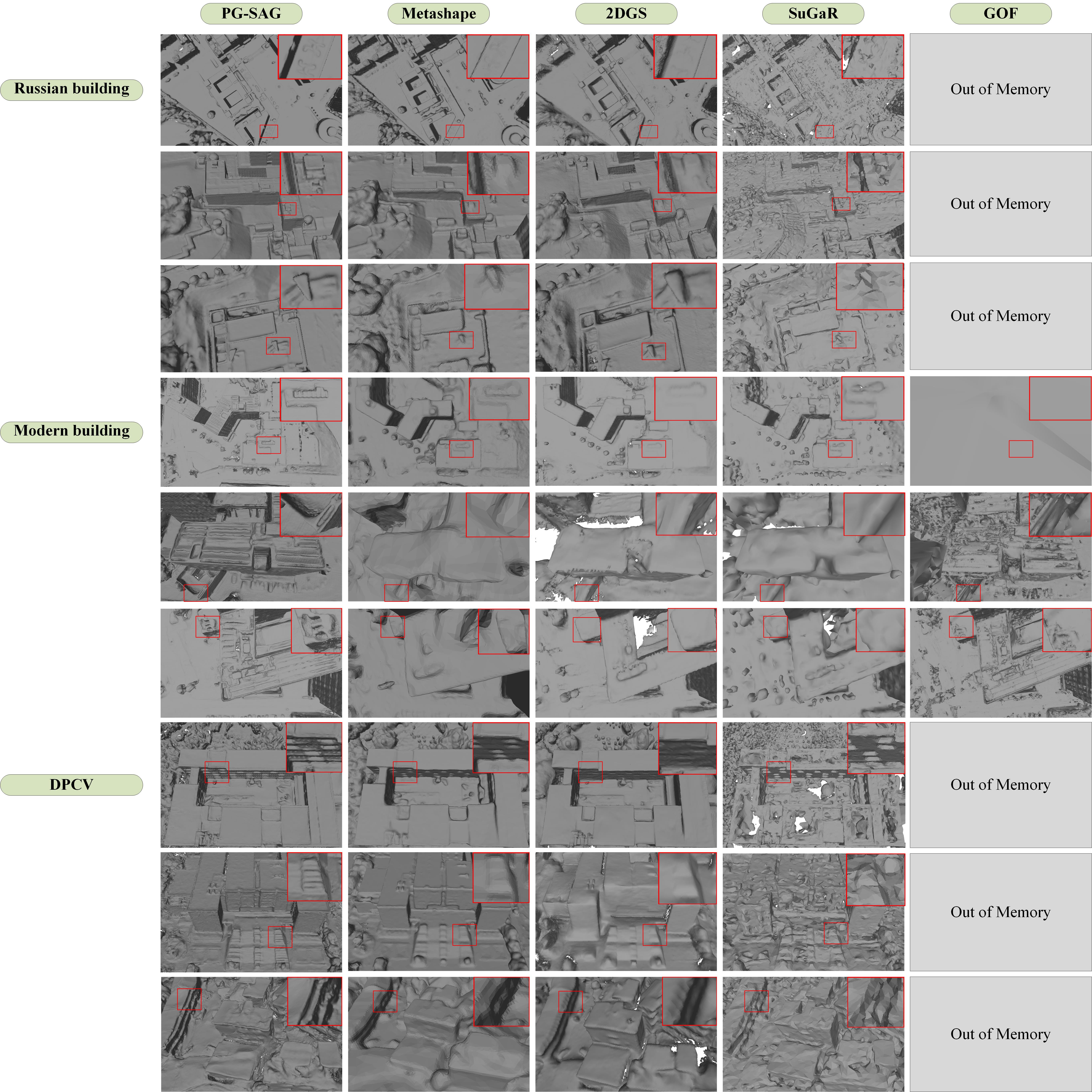}
\caption{Comparison of mesh reconstruction results across the Russian building, Modern building, and DPCV datasets with various methods. Notably, the GOF method fails to extract meshes for the Russian building and DPCV datasets due to memory limitations. For clarity, each figure includes a zoomed-in view enclosed within the red bounding box, displayed in the upper-right corner, to provide a more detailed view of fine-grained reconstruction quality.}
\label{fig3}
\end{figure*}

\textbf{Quantitative evaluation.} Table.\ref{table:Gauu_scene} compares our method to the state-of-the-art methods regarding to the evaluation metrics of Precision, Recall, and F1 score on the GauU-Scene dataset (including Russian Building and Modern Building). For quantitative evaluation, the surface sampling point density and the downsampling density of the ground truth point cloud are both set to 0.35m, with a distance threshold of 0.6m. It can be seen that our PG-SAG significantly outperforms the other methods (including the commercial package - Metashape) on obtaining the best results of all metrics, meaning more accurate meshes are generated by employing our refined building masks, improved boundary loss and gradient constraint balance loss. The other methods have receded results, which might be stemmed from the noise of background and inaccurate normals of boundaries \cite{Sugar,2DGS}. This evaluation demonstrates the superiority of our PG-SAG in building mesh reconstruction of large-scale scene.  

\textbf{Qualitative evaluation.} Additionally, Fig.\ref{fig3} provides a detailed visual comparison of mesh reconstruction across various methods, including ours, Metashape, 2DGS, SuGaR, and GOF, applied to three diverse datasets: Russian building, Modern building, and DPCV. In general, the results explicitly show that our approach is much superior to the alternatives in terms of boundary preservation and overall reconstruction fidelity. In particular, more findings are:

For the Russian building dataset, our method effectively captures intricate architectural details, such as sharp corners and smooth facade transitions, which are either blurred or distorted shown by all the other methods. Notably, the GOF method fails entirely due to memory limitations. While Metashape and 2DGS produce smoother but less precise reconstructions, SuGaR introduces significant distortions and noise, particularly in finer details.

In the Modern building dataset, the propsoed \textbf{\textit{PG-SAG}} again exhibits superior reconstruction results, especially in preserving the geometric integrity of structural boundaries and fine-grained features such as roof contours. In contrast, the compared methods introduce various artifacts, including unnatural bulges and interruptions in edge continuity. Furthermore, some methods fail to delineate key structural regions accurately, highlighting the robustness of our method in capturing complex geometries.

For the DPCV dataset that presents extra challenges due to its dense and complex building layouts (as seen in Fig. \ref{sample_images}), our method excels in managing overlapping structures while maintaining boundary clarity. The other methods degenerate on pronounced boundary distortions, incomplete reconstructions, or overly smoothed outputs that compromise detail preservation. On the contrary, our results faithfully preserve both large-scale structural relationships and fine-grained details, ensuring a more accurate representation of the scene.

 The qualitative and quantitative findings presented underscore the strengths of our method in delivering high-fidelity reconstructions with exceptional boundary preservation and structural accuracy. Notably, our approach demonstrates outstanding performance in challenging scenarios where alternative methods often fail or yield suboptimal outcomes.

\subsection{Ablation Study}
To assess various aspects of the proposed \textbf{\textit{PG-SAG}}, we conducted several ablation experiments on the Russian Building dataset, including the test of using input images with various resolution, the effect of the proposed improved boundary-aware  normal loss and gradient-constrained balance loss.

\begin{figure}[htbp]
\centering
\includegraphics[width=0.5 \textwidth]{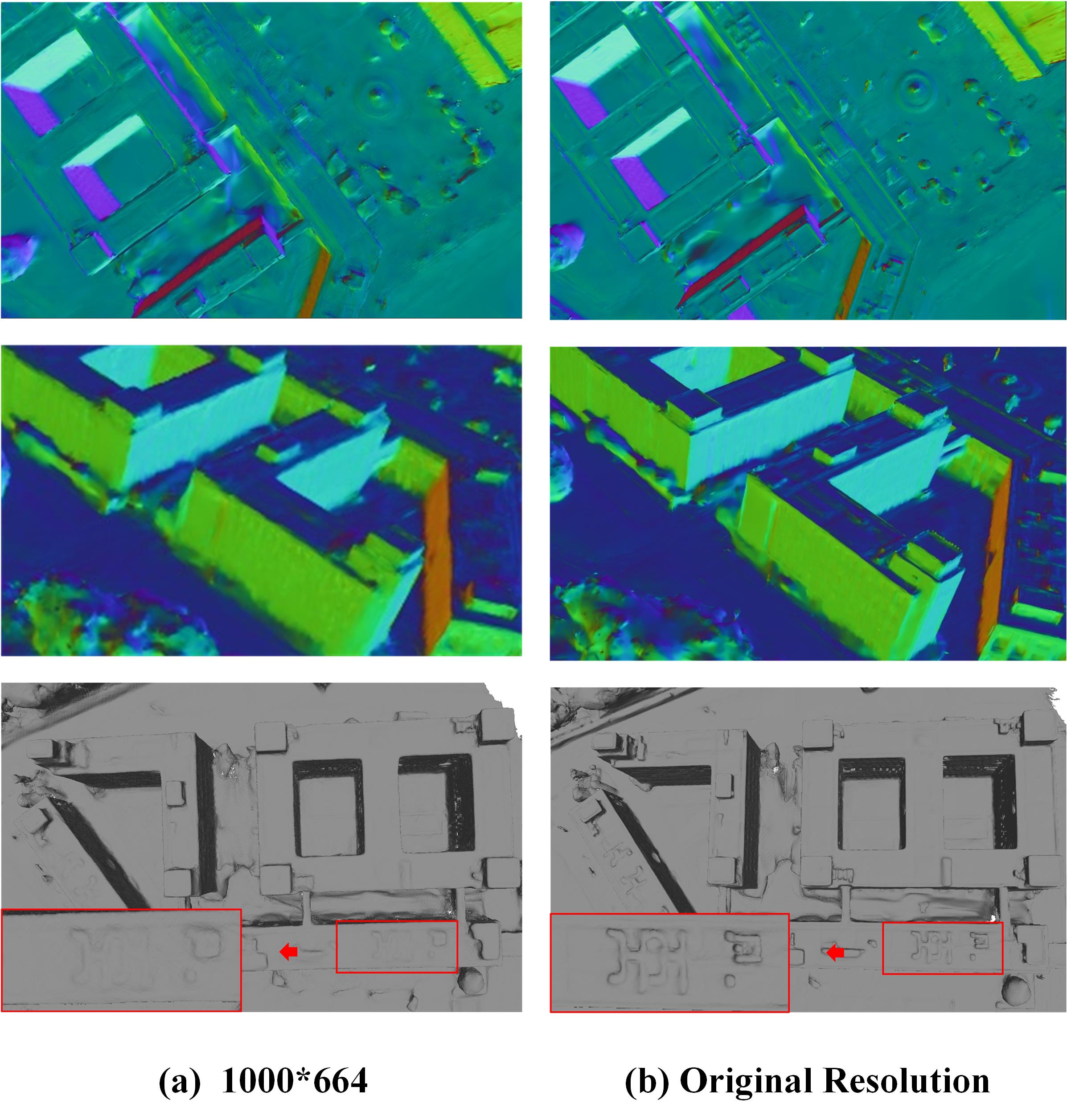}
\caption{Qualitative comparison of reconstruction results across two different resolutions. The first two rows depict the normal maps of the building surfaces from varying perspectives, while the final row showcases the corresponding mesh reconstruction outcomes.}
\label{fig4}
\end{figure}

\textbf{Resolutions. } As shown in Fig.\ref{fig4}, comparing to the results that from 1000*664 resoultion images (this is typically applied in many 3DGS-based methods, such as GOF\cite{GOF}, VastGaussian \cite{vastgaussian}), using high-resolution can generally yiled more fine-grained building surface, as using more information from the images is supported to provide more guidance during optimization. Thanks to the usage of refined building masks, we only need to compute a subset of pixels in each iteration which make the original resolution image accepted by our method. Furthermore, the geographical location of each group effectively controls the number and spatial distribution of Gaussian spheres. This enables our method to perform 3D reconstruction on high-resolution images, a capability that previous large-scale reconstruction methods could not tolerate.

\begin{figure}[htbp]
\centering
\includegraphics[width=0.5 \textwidth]{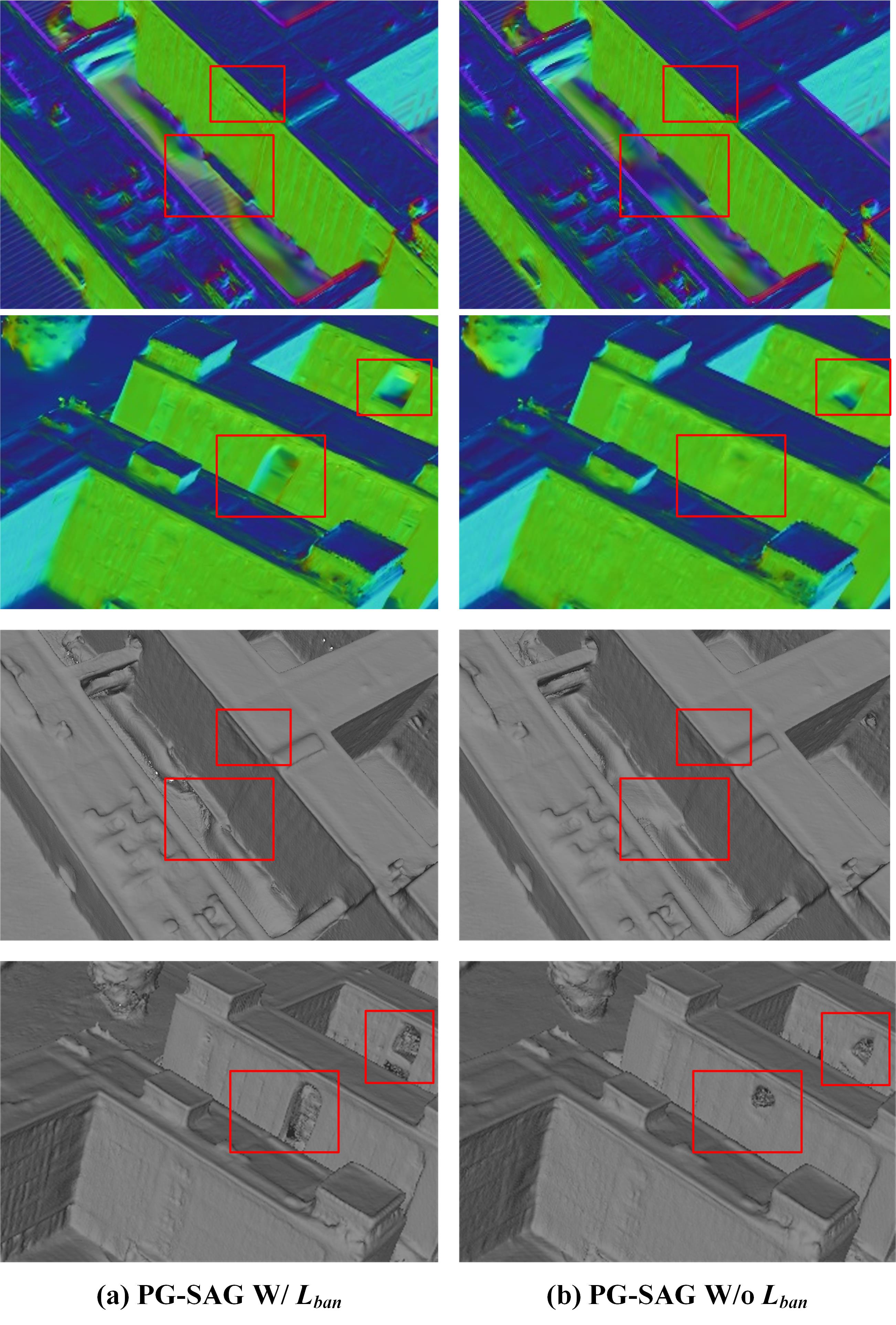}
\caption{Qualitative comparison of the proposed boundary-aware loss \( \mathcal{L}_{\text{ban}} \). The first two rows present the normal maps of the building surfaces from different perspectives, the last two rows illustrate the corresponding mesh results.}
\label{fig5}
\end{figure}

\textbf{Boundary-Aware Normal Loss.} Fig.\ref{fig5} illustrates the qualitative comparison of results by swithing on/off the proposed \( \mathcal{L}_{\text{ban}} \)(Boundary-Aware Normal Loss). The first two rows present normal maps of the building surfaces from different perspectives, while the last two rows display the reconstructed mesh results.

In boundary regions, due to the similarity in surface colors and textures of buildings, traditional methods often make flat surface assumptions, leading to inaccurate normal corrections. These errors are particularly clear in the red-boxed regions, where the normals deviate significantly without the \( \mathcal{L}_{\text{ban}} \), resulting in distorted or flattened mesh structures.

By integrating semantic information through \( \mathcal{L}_{\text{ban}} \), our method effectively distinguishes boundary regions and corrects the normals, as highlighted in the red boxes. This improvement ensures sharper boundary definitions, better structural preservation, and more accurate normal directions, which are crucial for high-quality mesh reconstruction. Consequently, the \( \mathcal{L}_{\text{ban}} \) contributes to significantly improved results in both the normal maps and the final mesh reconstructions, as evident in the comparison.

 \begin{table}[htbp]
\centering
\caption{Comparison of average training time for all building groups on Russian-building dataset.}
\label{table:time}
\begin{tabular}{lcccc}
\toprule
Methods  & Average time(min) \\
\midrule
Ours  & 26.3   \\
Ours without \( L_{\text{GC-load}} \) & 29.4 \\
Ours without \( L_{\text{ban}} \) & 25.2   \\
\bottomrule
\end{tabular}
\end{table}

\begin{figure}[htbp]
\centering
\includegraphics[width=0.5 \textwidth]{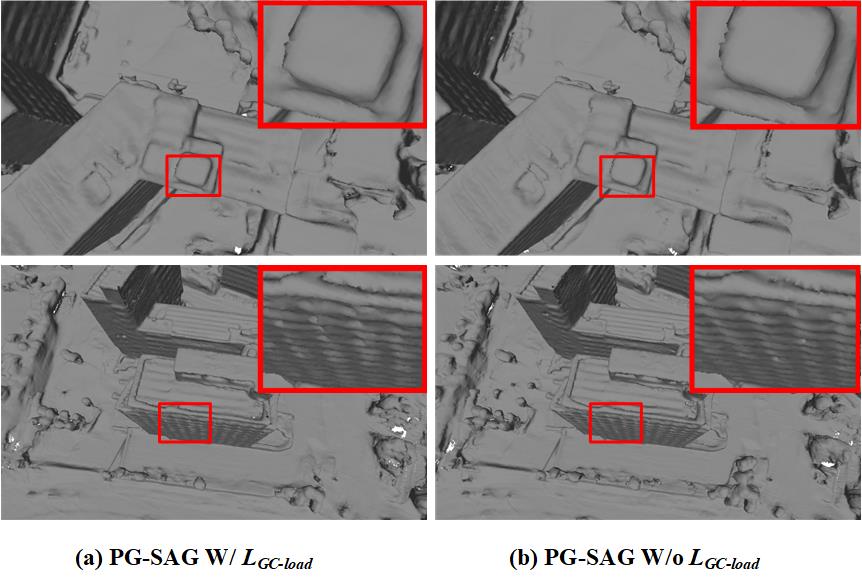}
\caption{Qualitative comparison of the proposed Gradient-constrained Balance-load Loss  \( \mathcal{L}_{\text{GC-load}} \).}
\label{load}
\end{figure}

\textbf{Gradient-constrained Balance-load Loss.} As section \ref{GCBL} explains, the presented Gradient-constrained Balance-load is expected to improve training time while minimizing the reduction of mesh quality. Table.\ref{table:time} lists the averaging training time for all building groups on Russian-building dataset, it can be see that the use of Gradient-constrained Balance-load Loss is capable to reduce the average training time for each building on the russian dataset by 12\%, because it adjusts the number of Gaussian spheres according to the complexity of the scene and reduces the waiting time of different threads. On the other hand, it is slightly slower when incorporating \( \mathcal{L}_{\text{ban}} \) which need extra efforts to deal with boundary information. Fig.\ref{load} compares the mesh results of with/without Gradient-constrained Balance-load Loss, it can be found that using \( \mathcal{L}_{\text{GC-load}} \) creates only very tinny influences on the final meshes.

\section{Conclusion and Limitation}

In this work, we introduce PG-SAG, a novel 3DGS-based method for fine-grained building reconstruction in large-scale urban scenes. By leveraging a semantic-aware grouping strategy, PG-SAG efficiently manages computational constraints, allowing for high-resolution image processing without the need for downsampling. Our method addresses key challenges in urban building reconstruction, including boundary ambiguity and computational load, through the integration of boundary-aware normal loss and gradient-constrained balance-load loss. Experimental results demonstrate that PG-SAG not only improves the precision of building surface reconstruction but also reduces training time, making it a practical solution for large-scale urban applications. 
Although our method achieves accurate building masks, automatic segmentation models such as LSA demonstrate less effectiveness in identifying other types of features. In future work, we plan to incorporate depth information to enhance semantic and geometric constraints for more comprehensive segmentation and reconstruction.

\section*{Declarations}


\paragraph{Acknowledgements}
 I would like to thank Butian Xiong for the GauU-Scene dataset.

\paragraph{Funding}
This work was supported by the National Natural Science Foundation of China (No.42301507) and Natural Science Foundation of Hubei Province, China (No. 2022CFB727).

\paragraph{Conflicts of interest/Competing interests}
Tengfei Wang, Xin Wang, Yongmao Hou, Yiwei Xu, Wendi Zhang and Zongqian Zhan declare that they have no competing interests.

\paragraph{Availability of data and material}
The authors have no permission to share these datasets
.

\paragraph{Code availability}
Our code is available on the website: \url{https://github.com/TFWang-9527/PG-SAG}.

\paragraph{Authors' contributions}
All the authors have contributed substantially to this manuscript. Conceptualization, Zongqian Zhan, Xin Wang, Tengfei Wang; methodology, Tengfei Wang, Xin Wang; formal analysis, Tengfei; investigation, Tengfei Wang and Zongqian Zhan; Visualization, Tengfei Wang and Wendi Zhang; Data Curation, Yongmao Hou and Yiwei Xu; writing-original draft preparation, Tengfei Wang and Xin Wang; supervision, Zongqian Zhan and Xin Wang; project administration, Zongqian Zhan and Xin Wang; funding acquisition,  Zongqian Zhan and Xin Wang. All authors have read and agreed to the published version of the manuscript.

\bibliographystyle{spbasic}      
\bibliography{bibliography}   

\end{document}